\documentclass[11pt]{article}
\usepackage[margin=1in]{geometry}
\usepackage[round,authoryear]{natbib}

\usepackage{amsmath,amssymb,amsfonts,amsthm,mathtools}
\usepackage{mathrsfs}
\usepackage{graphicx}
\usepackage{booktabs}
\usepackage{xcolor}
\usepackage{microtype}
\usepackage{hyperref}
\usepackage{url}
\usepackage[english]{babel}
\usepackage{comment}
\usepackage{authblk}
\hypersetup{hidelinks}

\theoremstyle{remark}

\newcommand{\dd}{\mathrm d}

\title{Inverse design of implicit bias by symmetry, stochasticity and holomorphic parametrizations}

\author[1]{Nicola Aladrah}
\author[1]{Emanuele Ballarin}
\author[2]{Matteo Biagetti}
\author[2]{Alessio Ansuini}
\author[1]{Alberto d'Onofrio}
\author[1,3]{Fabio Anselmi\thanks{Corresponding author: \texttt{fabio.anselmi@units.it}}}

\affil[1]{Department of Mathematics, Informatics and Geosciences,
University of Trieste,
Via A. Valerio 12/1, 34127 Trieste, Italy}

\affil[2]{Area Science Park,
Padriciano 99, 34149 Trieste, Italy}

\affil[3]{McGovern Institute for Brain Research,
Massachusetts Institute of Technology,
77 Massachusetts Avenue, Cambridge, MA 02139, USA}

\date{}

\begin{document}
\maketitle

\begin{abstract}
Can we design a model such that its stochastic training favours a desired class of solutions without enforcing an explicit penalty? Under suitable conditions, the interplay between symmetries of a model's
weight parametrization and stochastic training favours particular
solutions, inducing an implicit bias. Building on this mechanism,
we develop a framework for inverse-designing such biases by constructing novel
parametrizations and their associated symmetries. We show how holomorphic functions make this construction and calculation simple and explicit. Specifically, we introduce a new parametrization that biases learned weights toward
the binary values $\{-1,+1\}$. Numerical experiments confirm the theoretical predictions. They also show
that our parametrization reproduces the same preference induced by an
explicitly regularized model without adding a penalty to the training loss.
\end{abstract}

\section{Introduction}
\label{sec:introduction}

Modern overparametrized learning systems can fit their training data in many
ways. The loss function considers all interpolating solutions as equivalent,
but the training procedure does not. The model parametrization, initialization, learning rate, mini-batch sampling,
and noise during training influence which predictor
the optimizer converges to. This selection mechanism, usually called implicit bias
or implicit regularization, helps explain why predictors with similar
training error can have different structural and generalization properties
\citep{Vardi2023}.

Most studies of implicit bias address the following question: `Given a model
and an optimizer, which solutions does training favour?'
Here, we study the reverse question: can we design a parametrization
such that stochastic training favours prescribed predictors,
thereby inducing a regularization without adding an explicit penalty term to the loss? We refer to this problem as
\emph{inverse design of implicit bias}. 
Thus, our aim is to encode the desired predictor preference in the
geometry of the parametrization and let the optimizer reach it.

The key idea is that different parameter values can specify the same
predictor. A predictor-preserving symmetry is a continuous transformation
of the parameters that leaves the model's predictions unchanged.
Parameters can therefore vary along symmetry directions without changing the function represented by the model. When the infinitesimal symmetry generator is the gradient of a scalar
function of the parameters, continuous-time gradient flow preserves
that function, which we call a \emph{charge} by analogy with conserved
quantities in theoretical physics.
A finite optimization
step can break this conservation law because, although its first-order contribution vanishes by symmetry, higher-order terms do not. We will show that, under suitable conditions, repeated stochastic optimization steps
can drive the charge toward a specific value, thus biasing training
toward particular predictors among those that fit the data.

Charge conservation, finite-step breaking of conservation laws, and stochastic
selection have been studied from different perspectives and with
different objectives by \citet{kunin,tanaka}.
Noise-induced selection has also been investigated by
\citet{ziyin2024,balance}. We start from these ideas and results to introduce the key mathematical
objects needed to formulate and address the problem of inverse design
of implicit bias.

\paragraph{Contributions.} Our first and foremost contribution is a constructive approach to the inverse design of implicit biases under stochastic training, using holomorphic functions. The Cauchy--Riemann relations provide a simple way to construct
predictor-preserving symmetries and their associated charges.
At the critical points of the holomorphic function, the symmetry
generator vanishes, so the corresponding parameter values remain
unchanged under the symmetry transformation.
Under suitable conditions, we can enforce these points to map to targeted predictor values.

Our second contribution connects geometry to stochastic training dynamics.
We derive a finite-step update for the charge and establish
a sufficient condition on the charge second moment that ensures its convergence to a prescribed level under stochastic training.

Our third contribution is an explicit calculation of the bias
on the selected charge level, under the assumption that
the predictor evolves on a much
faster timescale than the charge.

Finally, our simulations provide a controlled test for the proposed mechanism.
We introduce a novel parametrization that biases learned weights
toward the binary values $\{-1,+1\}$. The performance of the new model is on a par with that of a tuned,
explicitly regularized baseline.

Together, these contributions provide a framework for the inverse design
of implicit bias, with the potential to shape which solutions models
learn and, consequently, their generalization properties.

\section{Related Work}
\label{sec:related-work}

\paragraph{Implicit bias and optimization.}
Foundational results connect gradient descent to maximum-margin
classification on separable data \citep{Soudry2018}, relate solution
selection to optimization geometry \citep{Gunasekar2018Geometry}, and
characterize function-space margins for gradient flow in infinitely
wide two-layer networks with homogeneous activations
\citep{ChizatBach2020}. 
More recent work extends these analyses to logistic regression at the
edge of stability \citep{Wu2023ImplicitLR} and to broader families of
multi-class losses \citep{Ravi2024ImplicitSeparable}.
For non-homogeneous networks, \citet{Cai2025NonHomogeneous} establish 
convergence under an exponential loss to a
margin-maximization problem. 

\paragraph{Parametrizations and implicit bias.}
Mirror descent provides a way to control implicit bias through
the choice of mirror potential \citep{sun2023unified}.
Related work connects continuous mirror descent to gradient flow
under suitable reparametrizations
\citep{amid2020reparameterizing}, including a characterization
for commuting parametrizations \citep{li2022implicit}.
These results provide a complementary route to controlling
implicit bias through optimization geometry. 
Additional related work characterizes conservation laws of gradient flows
using the Lie algebra generated by the model's Jacobian fields
\citep{marcotte2023abide}.

\paragraph{Geometry and stochasticity.}
The structure of gradient noise during training
can shape the solutions selected by stochastic gradient descent (SGD).
\citet{Zhu2019Anisotropic} relate escape from sharp minima to the
alignment between noise covariance and loss curvature, while
\citet{xie2021a} develop a diffusion theory of flat-minimum selection.
\citet{Wu2022Alignment} connect noise alignment to linear stability
and bounds on the sharpness of accessible minima in square-loss models.
Additional work describes landscape-dependent noise
\citep{FengTu2021} and effective regularization favouring flat solutions
\citep{Yang2023Landscape}. Recent work also examines the relation between noise covariance and the Hessian
\citep{Zhang2026NoiseCurvature}.
More broadly, noise covariance shapes implicit bias
\citep{haochen2021shape,Li2021WhatHappensAfterSGD}.
Label noise also induces implicit regularization favouring
flatter solutions
\citep{Blanc2020LabelNoise,Damian2021LabelNoise}.
For diagonal and quadratic parametrizations (called \textit{Hadamard parametrizations}), stochasticity can promote
sparse solutions \citep{pesme2021implicit,pillaudvivien2022labelnoise},
while finite-step effects also alter the bias \citep{even}.
Parameter symmetries provide another perspective on stochastic
selection through noise-induced equilibration and noise balance
\citep{ziyin2024,balance}. We adopt this latter perspective in our paper.
Related work derives an entropy formula from group-orbit
volumes in deep linear networks \citep{MenonYu2025Entropy}.
For Langevin dynamics with projected noise,
\citet{Menon2026ProjectedNoise} identify an implicit regularization
effect governed by orbit volume.

\paragraph{Singular learning theory.}
Singular learning theory studies how model singularities affect
Bayesian evidence and generalization
\citep{Watanabe2009,Watanabe2024Recent}. Our construction also involves singularities: the joint
predictor--charge Jacobian loses rank at designed critical points.

\section{Methodology}
\label{sec:method}
We begin with the well-known sparsity-promoting Hadamard parametrization
to introduce the main mathematical objects of our framework:
the predictor-preserving symmetry, its associated charge, and the joint
predictor--charge map. We examine how discrete stochastic training
affects the charge and induces an implicit bias. We derive the finite-step charge dynamics and give a sufficient
condition on the second charge moment for concentration on the zero-charge level. Under an explicit adiabatic assumption, discussed below, this
concentration selects a parameter set. The geometry of this set determines the
implicit bias, which we compute using the coarea formula \citep{federer1969gmt}.
This formula combines a change of variables with integration over
level sets, allowing us to express parameter volume in terms of
predictor and charge coordinates. The subsequent sections reverse the construction and use holomorphic functions to jointly design this charge and the predictor to favour targeted solutions.

\subsection{Hadamard parametrization}
\label{sec:hadamard}

Let $\Theta$ denote the parameter space and let $h:\Theta\to\mathcal{W}$ be the predictor map. Denote the predictor by $w=h(\theta)$. We consider the Hadamard parametrization: each predictor is
the product $w=u\odot v$, where $w_i=u_iv_i$. 
Since the parametrization acts
coordinatewise, we begin with the scalar predictor $w=h(u,v)=uv$.
For $w\neq0$, the set of parameters that gives the same predictor is
$h^{-1}(w)=\{(u,v)\in\mathbb R^2:uv=w\}$, i.e., a hyperbola. 

\begin{figure}[t]
  \centering
  \includegraphics[width=0.5\linewidth]{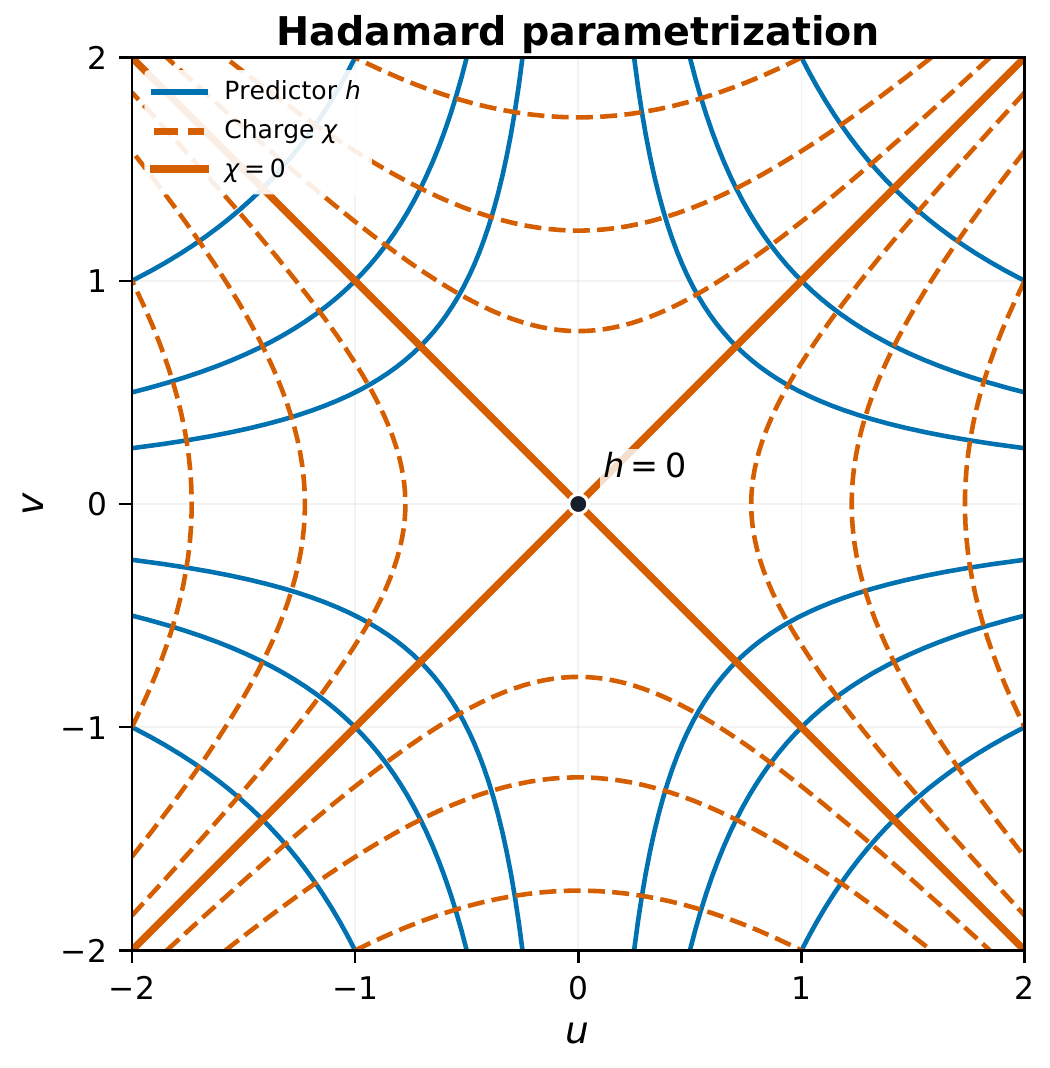}
  \caption{\textbf{Geometry of the Hadamard parametrization.}
  Predictor level sets $uv=\mathrm{const}$ (blue) and charge levels
  $\chi=(u^2-v^2)/2=\mathrm{const}$ (dashed orange) intersect
  orthogonally away from the origin. The zero-charge set consists
  of the balanced lines $u=\pm v$ (solid orange). The origin is
  the unique fixed point of the symmetry generator
  $\nabla\chi=(u,-v)$.}
  \label{fig:hadamard-level-sets}
\end{figure}

\paragraph{Symmetry and charge.}
The transformations $Q_\tau(u,v)=(e^\tau u,e^{-\tau}v)$ form a
one-parameter Lie group that preserves $h$. Its infinitesimal generator
$\xi(u,v)=\left.\partial_\tau Q_\tau(u,v)\right|_{\tau=0}=(u,-v)$
is tangent to each element in the set $h^{-1}(w)$. We suppose that the generator is the
gradient of a scalar function, the charge $\chi$.
This condition will be satisfied by definition in the holomorphic construction of Section~\ref{subsec:complex-construction}.
A simple calculation shows:
\begin{equation}
 \chi(u,v)=\frac{u^2-v^2}{2},
 \qquad \xi(u,v)=\nabla \chi(u,v),
 \qquad [\nabla h\cdot\nabla \chi](u,v)=0.
 \label{eq:hadamard-charge}
\end{equation}
Thus predictor and charge level sets intersect orthogonally
(Figure~\ref{fig:hadamard-level-sets}). In particular, for any batch $\mathcal B$, gradient flow on the sampled loss 
$L=\ell_{\mathcal B}(h)$ conserves $\chi$, since
$\dot \chi=-\ell_{\mathcal B}'(h)\nabla \chi\cdot\nabla h=0$
\citep{kunin,tanaka}.

In a discrete setting, the charge is no longer conserved but becomes a dynamical quantity.
Let $g_{\mathcal{B}}\equiv g_{\mathcal{B}_k}(w) = \partial_w \ell_{\mathcal{B}_k}(w)$ denote the loss gradient at $w$ for an independently sampled batch $\mathcal{B}$ and
$\theta=(u,v)$, where $k$ is the step number. A discrete SGD step is written as
$\theta_{k+1}=\theta_k-\eta g_\mathcal{B}\nabla h(\theta_k)$.
A Taylor expansion of the charge around the $k^{\text{th}}$-iterate gives
\begin{equation}
 \chi(\theta_{k+1})-\chi(\theta_k)
 =\frac{\eta^2g_\mathcal{B}^2}{2}
   \nabla h^{\mathsf T}\nabla^2\chi\nabla h +O(\eta^3).
 \label{eq:hadamard-curvature}
\end{equation}
The first-order term vanishes by symmetry since $\nabla\chi^{\mathsf T}\nabla h=0$, and, in the Hadamard case, there is no $O(\eta^3)$ term as $\chi$ is quadratic. Hence 
\begin{equation}
    \chi_{k+1}=(1-\eta^2g_\mathcal{B}^2)\chi_k.
    \label{eq:exact_chi}
\end{equation}
Thus, the charge dynamics appears at order $\eta^2$. The exact parameter-level derivation and the resulting
second-moment recursion are given in
Appendix~\ref{app:hadamard-M2}.

In order to calculate the implicit bias, we first have to determine where the charge concentrates. To do so, we analyse the dynamics of the charge second moment. Fix the predictor value $w$ and define the $j^\text{th}$-moment $\mu_j(w):=\mathbb E_{\mathcal B} [g_{\mathcal B}(w)^j]$ and the conditional charge law $\nu_w(\cdot, k)$. Then the charge second moment is 
\begin{equation}
    M_{2,w}(k)=\int_{\mathbb R} \chi^2 \; \nu_w(\dd \chi, k) = \mathbb E[\chi_k^2 \mid w].
\label{eq:charge_m_2}
\end{equation}
In the Hadamard case,
since the exact identity in equation~\eqref{eq:exact_chi} holds for every sampled batch, equation~\eqref{eq:charge_m_2} obeys a closed recursion. Squaring equation~\eqref{eq:exact_chi} and taking the expectation over the sampled batch gives
\begin{equation}
M_{2,w}(k+1)=\rho(w)\, M_{2,w}(k).
\label{eq:hadamard-M2-recursion}
\end{equation}
where
\begin{equation}
    \rho(w):=1-2\eta^2\mu_2(w)+\eta^4\mu_4(w)
=\mathbb E_{\mathcal B}\bigl[(1-\eta^2g_{\mathcal B}(w)^2)^2\bigr]
\end{equation}
Therefore, if $\rho(w)< 1$, then
\begin{equation}
    M_{2,w}(k) = \rho(w)^k M_{2,w}(0) \rightarrow 0, \qquad \text{when} \quad \eta^2\mu_4(w)<2\mu_2(w).
    \label{eq:condition}
\end{equation}
Consequently, $\nu_w(\cdot,k) \Rightarrow \delta(\chi)$, i.e. the
conditional charge law concentrates at $\chi=0$.

Appendix~\ref{app:hadamard-M2} derives this recursion and proves convergence of the conditional charge law. It also shows that, when $\mu_4(w)>0$, concentration around any centre $\chi_\star(w)$ necessarily implies $\chi_\star(w)=0$ for the Hadamard case. Thus, the Hadamard dynamics selects the balanced level $\chi=0$, where $u^2=v^2$.
For a general parametrization, equation~\eqref{eq:charge_m_2} need not yield
a closed recursion for $M_{2,w}(k)$, whose convergence must therefore
be established case by case.   

\paragraph{Adiabatic description and implicit bias.}
\label{subsec:charge-to-coarea}
As discussed above, a discrete SGD step changes the predictor at order $O(\eta)$,
whereas the charge changes at order $O(\eta^2)$ (see Appendix~\ref{app:hadamard-M2}).
This motivates an adiabatic assumption: the charge evolves on a much
slower timescale than the predictor.

For the Hadamard parametrization,
condition~\eqref{eq:condition} implies that the charge concentrates
at the zero-charge level. The selected support is
\begin{equation}
  \mathcal S_w=\{\theta:h(\theta)=w,\ \chi(\theta)=0\}.
  \label{eq:selected_supp}
\end{equation}
Any parameter law at fixed $w$, if it exists, is therefore supported on~\eqref{eq:selected_supp}.
Figure~\ref{fig:hadamard-level-sets} illustrates the
selection at fixed $w\ne0$. The predictor level set
$h=w$ (blue curves) intersects the selected charge
level $\chi=0$ (solid orange lines) at exactly two
points. These points form $\mathcal S_w$. The other
charge levels (dashed orange curves) illustrate what
is excluded dynamically as the charge concentrates at zero level.
We use the coarea formula \citep{federer1969gmt} to quantify the
parameter volume associated with each predictor value $w$ on the
selected charge level $\chi=0$. This weights each parameter
representation by the inverse Jacobian determinant of the joint map
$\Psi(\theta)=(h(\theta),\chi(\theta))$.
Selecting the charge level is expressed by the distributional measure
$\delta(\chi(\theta))\,\mathrm d\theta$.
For a test function $f$ supported at regular
predictor values, the coarea identity gives
\begin{equation}
      \int_{\Theta}
    f(h(\theta))
    \delta(\chi(\theta))\,\mathrm d\theta
  =
  \int_{\mathbb R}
    f(w)\Gamma_0(w)\,\mathrm dw.
\end{equation}
The resulting weight is
\begin{equation}
  \Gamma_0(w)
  =
  \sum_{\theta\in\mathcal S_w}
    \frac{1}{|\det D\Psi(\theta)|},
  \label{eq:intro_weight}
\end{equation}
for regular $(w,0)$. As mentioned above, coarea assigns each selected point the inverse-Jacobian
weight appearing in equation~\eqref{eq:intro_weight}. We interpret the corresponding quantity as a form of \textit{implicit bias}
\begin{equation}
  L_{\mathrm{IB}}(w)=-\log\Gamma_0(w).
  \label{eq:general-coarea-factor}
\end{equation}
Indeed, for equal loss values, a larger $\Gamma_0(w)$ represents more parameter volume on the selected charge level and hence a geometric preference for $w$.
Full details are given in
Appendix~\ref{app:coarea-derivation}.

\paragraph{Hadamard implicit bias.}
\label{subsec:hadamard-ib}
For the Hadamard parametrization $\chi(u,v)=\tfrac12(u^2-v^2)$. Fix $w\neq0$. Charge contraction selects $\chi=0$, so the selected parameter values satisfy $uv=w$ and $u^2=v^2$. There are exactly two solutions to those constraints. The joint Jacobian is
\begin{equation}
    D\Psi(u,v)
  =
  \begin{pmatrix}v&u\\u&-v\end{pmatrix},
  \qquad
  |\det D\Psi(u,v)|=u^2+v^2.
\end{equation}
On the selected level $u^2+v^2=2|w|$.
Equation~\eqref{eq:intro_weight} contains therefore the sum of two inverse
Jacobians, each equal to $1/(2|w|)$:
\begin{equation}
    \Gamma_0(w) = \frac{1}{|w|}.
\end{equation}
The reasoning can be easily extended to multiple coordinates, leading to
\begin{equation}
\Gamma_0(w)=\prod_{i=1}^{d}|w_i|^{-1},
  \qquad
  L_{\mathrm{IB}}(w)=\sum_{i=1}^{d}\log|w_i|.
  \label{eq:hadamard-log-bias}
\end{equation}
This example links a predictor-preserving symmetry, a dynamically selected
charge level, and implicit bias. 
We now see how these mathematical objects can be inverse designed using holomorphic functions to target specific $\Gamma_0$ .

\subsection{Inverse design of the implicit bias by holomorphic functions}
\label{subsec:complex-construction}
Given a finite collection of target predictor values, we seek a
parametrization whose induced implicit bias favours them.
Holomorphic functions provide such a construction, since their real
and imaginary parts have orthogonal gradients, $\nabla\chi^\top \nabla h=0$. We can therefore
identify the former with the charge and the latter with the predictor.
Importantly, as shown below, by designing the critical points of the holomorphic
function and their associated predictor values, we can shape the bias
toward the prescribed targets.

\paragraph{From holomorphic functions to symmetries, charges, and the implicit bias.}
Let the function
\begin{equation}
    F(\zeta)=A(u,v)+\mathrm{i}B(u,v),\qquad \zeta = u+\mathrm{i}v
    \label{eq:holo}
\end{equation}
be holomorphic. We define the charge and predictor by $\chi(u,v)=A(u,v)$ and $h(u,v)=B(u,v)$.
The Cauchy--Riemann relations,
$\partial_u A=\partial_v B$ and $\partial_v A = -\partial_u B$, imply
\begin{equation}
  \nabla \chi^{\mathsf T}\nabla h=0.
  \label{eq:cr-identities}
\end{equation}
Consequently, $\nabla\chi$ generates a local
predictor-preserving transformation. It vanishes exactly
at the critical points $F'(\zeta)=0$.
Crucially, note that
\begin{equation}
|F'(\zeta)|^2
=
(\partial_u A)^2+(\partial_v A)^2
=
\|\nabla\chi\|_2^2
=
|\det D\Psi|
\label{eq:holomorphic-jacobian}
\end{equation}
where the Jacobian is that of the joint map $\Psi=(h,\chi)$.
Thus, each regular representative on $\chi=0$ contributes $1/|F'(\zeta)|^2$ to $\Gamma_0(w)$ in equation~\eqref{eq:intro_weight}. By placing critical
points $F'(\zeta_j)=0$ on the dynamically selected charge level and their associated
predictor values through $h$, we can design where the induced
implicit bias is concentrated.
Appendix~\ref{app:local-lie-symmetry} gives
the full calculation.

Thus, the final step in the inverse design is to construct a holomorphic function with prescribed critical points. 
For a target polynomial holomorphic function $F=P$, we prescribe its critical points 
$\zeta_0,\ldots,\zeta_{K-1}$ by inverse design the zeros of its derivative
\begin{equation}
  P'(\zeta)
  =\alpha\prod_{j=0}^{K-1}(\zeta-\zeta_j),
  \qquad \alpha\ne0
  \label{eq:designed-polynomial-derivative}
\end{equation}
and then integrating to obtain $P$.

As an example, we recover the well-known sparsity-promoting Hadamard
parametrization within this framework. To favour zero predictors,
the minimal choice is $P'(\zeta)=\zeta$, which integrates to
$P(\zeta)=\zeta^2/2$. Taking its real and imaginary parts gives
$\chi(u,v)=(u^2-v^2)/2$ and $h(u,v)=uv$, recovering equations~\eqref{eq:hadamard-charge}.

\subsection{Inverse design: binary quantization}
\label{sec:binary}

For the target dictionary $w_\star=\{-1,+1\}$, we seek a
polynomial $P$ whose critical points $\zeta_\pm$ lie on the zero-charge level
and map to the predictor values $w=\pm1$. Following equation~\eqref{eq:holo} we define 
$\chi=\operatorname{Re}P$ and $h=\operatorname{Im}P/2$. These
requirements amount to
\begin{equation}
P'(\zeta_\pm)=0,
\qquad
P(\zeta_\pm)=\pm2\mathrm{i}.
\end{equation}
The factor $1/|P'(\zeta)|^2$ then produces the desired singular
bias at the target values, as follows from equations~\eqref{eq:intro_weight} and~\eqref{eq:holomorphic-jacobian}.
To construct such a polynomial, we choose the critical points
at $\zeta_\pm=\pm\mathrm{i}$ and take the simplest derivative
with these zeros,
\begin{equation}
P'(\zeta)=a(\zeta-\mathrm{i})(\zeta+\mathrm{i}).
\end{equation}
Integrating with zero constant gives
$P(\zeta)=a(\zeta^3/3+\zeta)$, so that
$P(\pm\mathrm{i})=\pm2a\mathrm{i}/3$.
The prescribed critical values therefore fix $a=3$, yielding
\begin{equation}
P'(\zeta)=3(\zeta-\mathrm{i})(\zeta+\mathrm{i}),
\qquad
P(\zeta)=\zeta^3+3\zeta.
\end{equation}
Taking the real and imaginary parts of $P$ yields the wanted parametrization $h$ and charge $\chi$:
\begin{equation}
  \chi(u,v)=u(u^2-3v^2+3),
  \qquad
  h(u,v)=\frac12v(3u^2-v^2+3).
  \label{eq:binary-pair}
\end{equation}
Figure~\ref{fig:binary-overview} (a) illustrates the corresponding predictor--charge geometry and the critical representatives \((0,\pm1)\). More generally, imposing simultaneously \(\chi=0\) and \(h=w=\pm1\) is equivalent to
$P(\zeta)=2\mathrm{i}w$. This specializes the set in equation~\eqref{eq:selected_supp} to this parametrization: $\mathcal S_w = \{\zeta: P(\zeta)=2\mathrm{i}w\}$ . Using equation~\eqref{eq:intro_weight}, where \(|\det D\Psi(\chi,h)|=\tfrac12|P'(\zeta)|^2\), we have:
\begin{equation}
  \Gamma_0(w)
  =\sum_{\zeta:\,P(\zeta)=2\mathrm{i}w}
    \frac{2}{|P'(\zeta)|^2},
  \qquad w\notin\{-1,+1\}.
  \label{eq:binary-joint-jacobian}
\end{equation}
To characterize the implicit bias near each target value
$w_\star\in\{-1,+1\}$, we expand $P$ around the corresponding
critical point $\zeta=\mathrm{i}w_\star$. This yields
\begin{equation}
  \Gamma_0(w)\sim\frac{1}{6|w-w_\star|},
  \qquad
  L_{\mathrm{IB}}(w)=\log|w-w_\star|+O(1)
  \quad\text{as }w\to w_\star.
  \label{eq:binary-local-bias}
\end{equation}
Appendix~\ref{app:binary-analysis} gives the full derivation.

\paragraph{Discrete dynamics and branch attraction.}
The calculations above complete the geometric part of the construction. The parametrization was chosen so that the critical points \((0,\pm1)\) make the Jacobian vanish, producing the singular behaviour of \(\Gamma_0(w)\) as \(w\to\pm1\). Thus the implicit bias identifies the prescribed binary values as the geometrically preferred predictors. It remains to see how the discrete stochastic dynamics actually drives the parameters toward the zero-charge branch containing these critical representatives.
Writing $g_{\mathcal B}$ for the sampled loss derivative,
a simultaneous SGD step gives
\begin{equation}
  \begin{aligned}
    u_{k+1}&=(1-3\eta v_k g_{\mathcal B})u_k,\\
    v_{k+1}&=v_k-\frac{3\eta g_{\mathcal B}}{2}
      (u_k^2-v_k^2+1).
  \end{aligned}
  \label{eq:binary-sgd}
\end{equation}
Under the sufficient negative-logarithmic-drift condition derived in
Appendix~\ref{app:binary-analysis}, the multiplicative update drives
\(u_k\to0\) and, for bounded \(v_k\), also gives \(\chi_k\to0\).
Hence the dynamics approaches the invariant branch \(u=0\).
On this branch, the remaining \(v\)-update vanishes at \(v=\pm1\),
so the corresponding critical representatives preserve the binary
predictor values \(h=\pm1\) under SGD
(see Appendix~\ref{app:binary-analysis}).

\section{Experiments}
\label{sec:experiment}

In the following, we test whether
the proposed parametrization reproduces the preference for binary
weights induced by explicit regularization, without adding a penalty
to the loss. We also track charge evolution to assess the proposed
selection mechanism.

\subsection{Recovering binary Ising interactions from spin energies}
\label{subsec:ising-recovery}

We apply a model with the binary parametrization described in
Section~\ref{sec:binary} to the problem of recovering signed pairwise
interactions in an Ising model \citep{Ising1925}.
The model is linear in the unknown interactions and provides
a controlled setting in which to examine charge evolution,
concentration near prescribed predictor values, and reconstruction accuracy.
We also measure the ratio of charge to predictor update norms as
a diagnostic of the timescale separation underlying the adiabatic assumption.
Our aim is to test how the proposed parametrization biases optimization
towards binary interactions without an explicit penalty, and to compare
its reconstruction performance with that of a tuned, explicitly
regularized model. We also include an unregularized vanilla model as a baseline.

\paragraph{Model, data and measurements.}
We consider a classical Ising model with energy
\[
E(s)=\sum_{i<j}J^*_{ij}s_i s_j
    =\frac12 s^{\mathsf T}J^*s,
\]
where $s\in\{-1,+1\}^{D}$ and the interaction matrix
$J^*\in\mathbb R^{D\times D}$ is symmetric with zero diagonal
and independent, uniformly sampled couplings
$J^*_{ij}\in\{-1,+1\}$ for $i<j$.
The factor $1/2$ ensures that each interaction is counted once.
Positive and negative couplings are therefore equally probable,
without enforcing an exact balance.
Our goal is to recover these couplings from a limited number of
energy observations. We use $D=60$, giving
$d=D(D-1)/2=1770$ independent unknowns.
For independently sampled uniform spin configurations $s_\mu$,
we observe the noiseless normalized energies
\begin{equation}
E_\mu
=\frac{1}{\sqrt d}\sum_{i<j}J^*_{ij}s_{\mu i}s_{\mu j}
=\frac{s_\mu^{\mathsf T}J^*s_\mu}{2\sqrt d},
\qquad \mu=1,\ldots,N.
\label{eq:ising-observations}
\end{equation}
These are measurements at random spin configurations, not samples
from an unknown equilibrium distribution.
Writing the ground-truth weights as
$w^*=(J^*_{ij})_{i<j}\in\mathbb R^d$, we construct
$X\in\mathbb R^{N\times d}$ with one column for each pair $i<j$:
\[
X_{\mu,(i,j)}=\frac{s_{\mu i}s_{\mu j}}{\sqrt d}.
\]
Thus, each row $X_\mu$ contains $d$ features and
$E_\mu=X_\mu w^*$.
We use $N=1062$ observations for training and evaluate each model
on $2048$ additional independent spin configurations with the
same interaction matrix.

\begin{figure}[tbp]
  \centering
  \includegraphics[width=\linewidth]{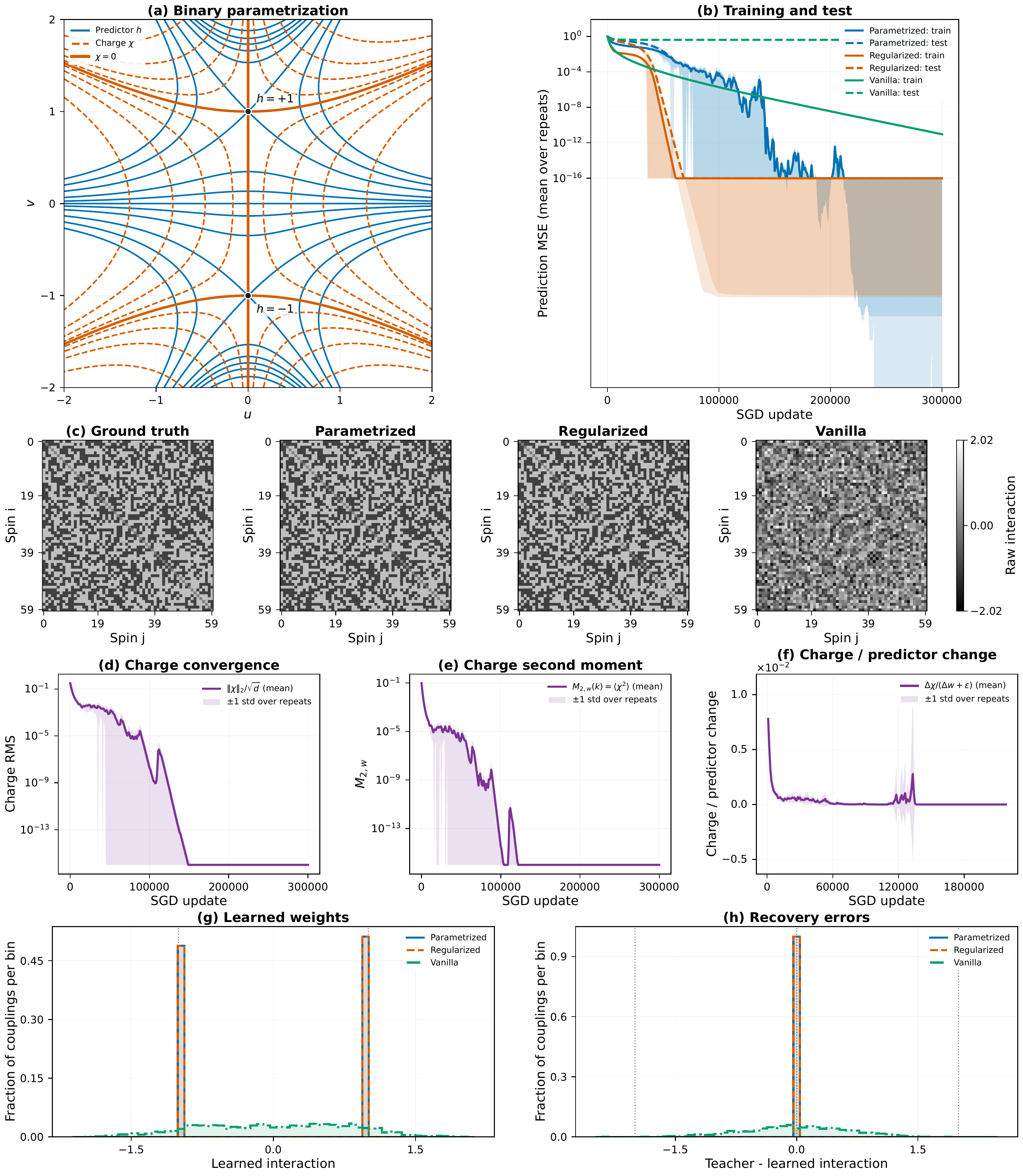}
  \caption{\textbf{Binary Ising interaction recovery with three independently
  tuned methods.}
  \textbf{Top row: (a)} Predictor level sets (blue), charge level sets
  (dashed orange), and the zero-charge set (thick orange); the marked
  critical points yield $h=\pm1$.
  \textbf{(b)} Prediction MSE for training (solid) and
  test (dashed) for the parametrized (blue), regularized
  (orange), and vanilla (green) models.
  \textbf{Second row: (c)} Ground-truth, parametrized, regularized,
  and vanilla interaction matrices from the first run, using a shared
  colour scale (all runs gave qualitatively similar results).
  \textbf{Third row: (d)} Charge RMS
  $\|\boldsymbol \chi\|_2/\sqrt d$ for the parametrized model.
  \textbf{(e)} Charge second moment $M_{2,w}=\langle\boldsymbol\chi^2\rangle$ for the parametrized model.
  \textbf{(f)} Charge-to-predictor change ratio for the parametrized model.
  \textbf{Bottom row: (g)} Distributions of raw learned couplings for the same run as in (c).
  \textbf{(h)} Distributions of true-minus-learned coupling errors for the same run as in (c).}
  \label{fig:ising-overview}
  \label{fig:binary-overview}
\end{figure}

\paragraph{Parametrized, regularized and vanilla training.}
The parametrized model applies the map $h$ in
\eqref{eq:binary-pair} independently to each coupling, setting
$\widehat J_{ij}=h(u_{ij},v_{ij})$ for $i<j$ and enforcing symmetry
and a zero diagonal. An explicitly regularized baseline optimizes
these couplings directly with a double-well penalty that favours
the same binary values. Finally, a vanilla baseline optimizes
couplings directly, without any penalty.
For minibatch $\mathcal B$, the objectives are
\begin{equation}
\begin{aligned}
&L^{\mathrm{param}}(u,v)
 = \frac{1}{2|\mathcal B|}
    \sum_{\mu\in \mathcal B}
   \bigl(X_\mu h(u,v)-E_\mu-\xi_{\mu,k}^{\mathrm{param}}\bigr)^2,\\
&L^{\mathrm{reg}}(w)
 = \frac{1}{2|\mathcal B|}
    \sum_{\mu\in \mathcal B}
   \bigl(X_\mu w-E_\mu\bigr)^2
   +\frac{\lambda}{d}\sum_{e=1}^d(w_e^2-1)^2,\\
&L^{\mathrm{van}}(w)
 = \frac{1}{2|\mathcal B|}
    \sum_{\mu\in \mathcal B}
   \bigl(X_\mu w-E_\mu\bigr)^2.
\end{aligned}
\label{eq:ising-training-loss}
\end{equation}
Here, $h(u,v)$ is applied component-wise.
For the parametrized model, independently resampled Gaussian noise
is added to the energy targets at each training iteration, with
$\xi_{\mu,k}^{\mathrm{param}}\sim
\mathcal N(0,\sigma_{\mathrm{param}}^2)$.
This training noise sustains stochastic fluctuations near interpolation
and is intended to maintain the finite-step stochastic mechanism active
underlying the proposed charge-concentration mechanism (see Figure~\ref{fig:ising-overview}, (e)).
Neither baseline uses added target noise.
We set the batch size to $|\mathcal B|=1$ for all three models and sample
training indices uniformly with replacement at each iteration.
This choice yields good convergence while maximizing minibatch
sampling noise.

Each model is trained for $3\times10^5$ iterations using SGD.
For each coupling, we initialize
$u_{ij}\sim\mathcal N(0.1,0.02^2)$ independently and set $v_{ij}=0$,
giving $h(u_{ij},v_{ij})=0$.
Both baseline models start from the same effective weights.

We select hyperparameters for each method by
minimizing the final coupling reconstruction error $D_1=\sum_{i<j}|\widehat J_{ij}-J^*_{ij}|$
on a single initial run.
For the regularized model, we search
$\lambda\in\{0.01,0.1,0.3,0.6,1\}$ and
$\eta\in\{0.1,0.9,1.2\}$.
For the parametrized model, we search
$\eta\in\{0.1,0.6,0.9\}$ and
$\sigma_{\mathrm{param}}\in\{0.6,0.7,0.8\}$.
For vanilla regression, we search
$\eta\in\{0.1,0.6,0.9,1.2\}$.
The search grids were chosen through preliminary experiments covering
hyperparameters regimes with low $D_1$.
Candidates reuse the same initialization and
random seeds for sampling and training noise.
The selected settings are $(\eta,\lambda)=(1.2,0.1)$ for the
regularized model,
$(\eta,\sigma_{\mathrm{param}})=(0.6,0.6)$ for the parametrized model,
and $\eta=0.6$ for vanilla regression.
We then repeat the training procedure with these fixed selected settings 
across ten independent runs, varying the random seeds for stochastic training, 
while preserving the true interaction matrix, training data, and initialization.

\paragraph{Results.}
Across the ten runs, the parametrized and regularized models
achieve coupling reconstruction errors of
$D_1=(3.8\pm2.1)\times 10^{-16}$ and $D_1=(2.4\pm0.1)\times 10^{-12}$, respectively
(mean $\pm$ one sample standard deviation).
Vanilla regression instead achieves $D_1=892.96$ with a standard deviation below $0.01$.
The parametrized and regularized models therefore recover the
couplings much more accurately than vanilla regression, and achieve
near-exact numerical recovery across all runs.

The learning curves and coupling distributions in
Figure~\ref{fig:ising-overview} (b) and (g), respectively, compare prediction accuracy and
concentration near the binary targets of the learned weights.
The inverse-designed parametrization produces a preference for
binary values without an explicit penalty, whereas the regularized
baseline imposes this preference through its double-well penalty.\\
For the parametrized model, we track the charge RMS,
$\|\boldsymbol\chi_k\|_2/\sqrt d$.
We compare charge and predictor updates using $R=\langle\|\Delta\boldsymbol\chi\|_2\rangle
     /(\langle\|\Delta w\|_2\rangle+\varepsilon)
$, with $\varepsilon=10^{-12}$,
where the brackets denote averages over non-overlapping windows
of $1000$ SGD updates.

The underlying parameter trajectories are shown in
Figure~\ref{fig:binary-parameter-trajectories} in
Appendix~\ref{app:binary-analysis}: \(u\) approaches zero while
individual \(v\)-coordinates approach either \(+1\) or \(-1\),
consistent with the branch-attraction and critical-representative
dynamics derived in Appendix~\ref{app:binary-analysis}.

Curves and histograms in Figure~\ref{fig:binary-overview} show means
and bands showing one sample standard deviation across the ten independent
runs. The models share the same true interaction matrix, data, and
effective initialization across all runs, varying only the stochastic
training noise.
The displayed interaction matrices refer to the first run.

\section{Conclusions and Future Perspectives}
\label{sec:conclusions}

We introduced a constructive framework for designing implicit biases
through holomorphic parametrizations. The Cauchy--Riemann relations
connect predictor-preserving symmetries to conserved charges, while
the coarea formula characterizes the implicit bias induced by the
geometry of the charge level set selected by training. This framework recovers the known
sparsity-promoting Hadamard parametrization and provides the construction
of a novel parametrization that favours the learning of binary weights.
In a controlled numerical experiment, the latter reproduces the effect
of an explicit regularizer towards weight binarization without adding a penalty to the loss, thus
supporting the proposed mechanism for the inverse design of a chosen implicit bias.

A theoretical direction for future work is to move beyond
constructions that parametrize each predictor coordinate separately.
Allowing parameters to interact across coordinates could support
richer symmetry structures with multiple associated charges.
Developing such an extension requires identifying predictor-preserving
generators that guarantee closure relations under Lie brackets.
Further work could explore larger target dictionaries and more
complex models or datasets, potentially going beyond holomorphic
parametrizations, and using such theory to guide model design and training.
This work lays the theoretical foundations for such developments
by showing how implicit bias can be designed to shape which solutions
models learn, with potential implications for the control of
learned representations and ultimately generalization.

\subsection*{Reproducibility Statement}
Section~\ref{sec:method} defines the constructions and theoretical
assumptions; Appendices~\ref{app:local-lie-symmetry},
\ref{app:hadamard-M2}, \ref{app:coarea-derivation},
and~\ref{app:binary-analysis} provide detailed derivations and proofs.
Section~\ref{subsec:ising-recovery} specifies the synthetic data
generation, model parametrizations, training objectives,
initialization, hyperparameter selection, and evaluation protocol.
The experimental code is available in anonymized form at
\url{https://github.com/NicolaAdrah/inverse-design}.

\bibliographystyle{plainnat}
\bibliography{references}

\clearpage
\appendix

\section*{Appendix}

The supplementary material follows the logical construction used in the main text. Appendix~\ref{app:local-lie-symmetry} shows why harmonic conjugates (the real and imaginary part of a holomorphic function) generate predictor-preserving symmetries and relates their critical points to the implicit bias. Appendix~\ref{app:hadamard-M2} then studies the Hadamard parametrization, where the discrete charge dynamics closes exactly and proves concentration on the balanced level \(\chi=0\). Appendix~\ref{app:coarea-derivation} takes this dynamically selected level as input and derives the corresponding predictor-space implicit bias by coarea formula. Finally, Appendix~\ref{app:binary-analysis} applies the same geometric construction to the inverse-designed binary parametrization and separately studies whether discrete SGD is attracted toward the part of the zero-charge set containing the designed critical points.

\section{Harmonic conjugates generate local Lie symmetries}
\label{app:local-lie-symmetry}
The purpose of this section is to establish two properties used later in the inverse-design construction. First, the real part of a holomorphic function generates a direction along which its imaginary part is unchanged. Second, the same complex derivative controls the Jacobian entering the implicit bias.

Let $F(\zeta)=A(u,v)+\mathrm{i}B(u,v)$ be
holomorphic on an open domain
$\Omega\subset\mathbb C\simeq\mathbb R^2$,
where $\zeta=u+\mathrm{i}v$. Define
$\chi(u,v)=A(u,v)$, $h(u,v)=B(u,v)$, and $X=\nabla\chi$.
The Cauchy--Riemann equations give
$\partial_u B(u,v)=-\partial_v A(u,v)$ and
$\partial_v B(u,v)=\partial_u A(u,v)$. Therefore
\begin{equation}
\begin{aligned}
  X[h]
  &=
  \nabla A\cdot\nabla B\\
  &=
  (\partial_u A)(\partial_u B)
  +(\partial_v A)(\partial_v B)\\
  &=
  -(\partial_u A)(\partial_v A)
  +(\partial_v A)(\partial_u A)
  =0.
\end{aligned}
\end{equation}
Since $X$ is smooth, it generates a local flow
$\Phi_t$. Along this flow,
$$
  \frac{\mathrm d}{\mathrm dt}
  h(\Phi_t(\theta))
  =
  X[h](\Phi_t(\theta))
  =0.
$$
Thus $h\circ\Phi_t=h$ wherever the flow is
defined. Flow uniqueness gives
$\Phi_0=\mathrm{id}$,
$\Phi_t^{-1}=\Phi_{-t}$, and
$\Phi_t\circ\Phi_s=\Phi_{t+s}$ wherever these
expressions are defined. Hence $X$ generates
a local one-parameter Lie symmetry of the
predictor and of every loss depending only on it. Thus the same holomorphic function simultaneously provides the predictor \(h\) and a charge \(\chi\) whose gradient moves along predictor level sets. The next calculation shows why critical points of that function are also distinguished geometrically.\\
The same derivatives give the coarea Jacobian
used in the main text. Indeed,
$$
  F'(\zeta)
  =
  \partial_u A+\mathrm{i}\partial_u B
  =
  \partial_u A-\mathrm{i}\partial_v A,
$$
and, for $\Psi=(h,\chi)=(B,A)$,
\begin{equation}
\begin{aligned}
  |\det D\Psi|
  &=
  \left|
    (\partial_u B)(\partial_v A)
    -(\partial_v B)(\partial_u A)
  \right|\\
  &=
  (\partial_u A)^2+(\partial_v A)^2
  =
  |F'(\zeta)|^2.
\end{aligned}
\end{equation}
At regular points, each representative therefore
contributes $1/|F'(\zeta)|^2$ to the coarea
sum in equation~\eqref{eq:intro_weight}. If the predictor is rescaled by a constant factor, the joint Jacobian is rescaled by the same factor. This is the only modification needed in the binary example, where \(h=\operatorname{Im}P/2\).\\
For a coordinatewise construction on disjoint
parameter pairs, the generators $X_i$ preserve
every predictor coordinate and satisfy
$[X_i,X_j]=0$. Their commuting local flows
define a local action of the abelian Lie group
$(\mathbb R^d,+)$. Since
$|F'(\zeta)|^2=\|\nabla\chi(\zeta)\|_2^2$,
the fixed points of each local flow are precisely
the corresponding critical points of $F$.

\section{Hadamard charge dynamics}
\label{app:hadamard-M2}
This section establishes the dynamical selection of the Hadamard zero-charge level. We first derive the exact finite-step evolution of the charge. We then show that, after freezing the predictor according to the adiabatic approximation, its second moment contracts geometrically and the complete conditional charge law converges to \(\delta_0\). Finally, we show that no nonzero deterministic charge value can be a second-moment concentration center.

Set at step $k$, the predictor $w_k=u_kv_k$, the charge $\chi_k=\tfrac12(u_k^2-v_k^2)$ and the derivative for a minibatch $\mathcal B$, $g_{\mathcal{B}} =g_{\mathcal{B}_k}(w_k)$. One SGD step gives, $u_{k+1}=u_k-\eta g_{\mathcal{B}} v_k$ and $v_{k+1}=v_k - \eta g_{\mathcal{B}} u_k$. For the predictor,
\begin{align}
    w_{k+1} &= u_{k+1}v_{k+1}
    \nonumber\\
    &=(u_k-\eta g_{\mathcal{B}} v_k)(v_k - \eta g_{\mathcal{B}} u_k)
    \nonumber\\
    &= w_k - \eta g_{\mathcal{B}} (u_k^2+v_k^2) + \eta^2 g_{\mathcal{B}}^2 w_k.
\end{align}
And for the charge,
\begin{align}
    2\chi_{k+1} &= u_{k+1}^2 - v_{k+1}^2
    \nonumber \\
    &= (u_k-\eta g_{\mathcal{B}} v_k)^2 - (v_k - \eta g_{\mathcal{B}} u_k)^2
    \nonumber \\
    &= u_k^2-2\eta g_{\mathcal{B}} u_kv_k+\eta^2g_{\mathcal{B}}^2v_k^2
    -v_k^2+2\eta g_{\mathcal{B}} u_kv_k-\eta^2g_{\mathcal{B}}^2u_k^2
    \nonumber \\
    &= u_k^2 - v_k^2 - \eta^2 g_{\mathcal{B}}^2(u_k^2-v_k^2)
    \nonumber\\
    &=2(1-\eta^2 g_{\mathcal{B}}^2) \chi_k.
    \label{eq:chi_exact}
\end{align}
The \(O(\eta)\) terms cancel exactly, leaving a purely multiplicative \(O(\eta^2)\) change of the charge.
\subsection{Charge second moment}
The exact SGD process couples $w_k$ and $\chi_k$. Under the adiabatic assumption, we fix $w$, and let $(\mathcal B_k)_{k\geq0}$ be independently sampled minibatches and define, $A_k(w) = 1-\eta^2 g_{\mathcal{B}}^2$. The conditional charge dynamics is, $\chi_{k+1}=A_k(w) \chi_k$. Let $\nu_w(\cdot, k)$ be the law of $\chi_k$ under this conditioned process. Assume that $\nu_w(\cdot,0)$ has a finite second moment and that, $\mu_2(w)=\mathbb E_{\mathcal B}[g_{\mathcal B}(w)^2]<\infty$ and $\mu_4(w)=\mathbb E_{\mathcal B}[g_{\mathcal B}(w)^4]<\infty$. Define
\begin{equation}
    M_{2,w}(k) = \int_{\mathbb R} \chi^2 \nu_w(\dd \chi,k)
\end{equation}
Because $\mathcal B_k$ is sampled independently of the charge produced by the preceding minibatches,
\begin{align}
    M_{2,w}(k+1) &= \int_{\mathbb R} \mathbb E_{\mathcal B}[A_k(w)^2 \chi_k^2] \nu_w(\dd \chi,k)
    \nonumber\\
    &= \mathbb E_{\mathcal B}[A_k(w)^2] \int_{\mathbb R} \chi^2 \nu_w(\dd \chi,k)
    \nonumber\\
    &= \mathbb E_{\mathcal B}[(1-\eta^2 g_{\mathcal B}(w)^2)^2] M_{2,w}(k)
\end{align}
By expanding the square we recover equation~\eqref{eq:hadamard-M2-recursion}. Iterating the recursion gives,
\begin{equation}
    M_{2,w}(k) = \rho(w)^k M_{2,w}(0).
    \label{eq:geometric_recursion}
\end{equation}
where \(\rho(w)=1-2\eta^2\mu_2(w)+\eta^4\mu_4(w)\), therefore, if $0 \leq \rho(w) <1$, then $M_{2,w}(k) \to 0$.

\subsection{Convergence of the charge law}
The second-moment estimate has a direct probabilistic meaning: as \(k\) grows, the charge mass outside every fixed neighborhood of zero must vanish. For every $\varepsilon>0$,
\begin{equation}
  \nu_w(\{|\chi|\ge\varepsilon\},k)
  \le
  \frac{M_{2,w}(k)}{\varepsilon^2}
  =
  \frac{\rho(w)^k M_{2,w}(0)}{\varepsilon^2}
  \longrightarrow0.
\end{equation}
Thus the charge converges to zero in probability under
the frozen-predictor reduced process.
Let $f:\mathbb R\to\mathbb R$ be bounded and continuous. Given $\varepsilon>0$, continuity of $f$ at zero provides $\delta>0$ such that $|f(\chi)-f(0)| <\varepsilon$, whenever $|\chi|<\delta$. Then
\begin{align}
    \left| \int_{\mathbb R}f(\chi)\,\nu_w(d\chi,k)-f(0) \right| &\leq \int_{\{|\chi|<\delta\}} |f(\chi)-f(0)|\,\nu_w(d\chi,k)
    \nonumber\\
    &\quad+ \int_{\{|\chi|\geq\delta\}} |f(\chi)-f(0)|\,\nu_w(d\chi,k)
    \nonumber\\
    &\leq \varepsilon + 2\|f\|_\infty \nu_w\left( \{|\chi|\geq\delta\},k \right).
\end{align}
The second term converges to zero. Since $\varepsilon$ is arbitrary, $\int_{\mathbb R}f(\chi)\,\nu_w(d\chi,k) \longrightarrow f(0)$. Therefore
\begin{equation}
    \nu_w(\cdot,k) \Rightarrow \delta_0.
    \label{eq:nu_delta}
\end{equation}
For each fixed $w$, this identifies zero as the selected
charge value. The corresponding selected parameter set is
$\mathcal S_w=\{\theta:h(\theta)=w,\ \chi(\theta)=0\}$.
The support implication for any parameter law,
and the implicit bias assigned to this selected set, are
established in Appendix~\ref{app:coarea-derivation}.

\subsection{Concentration center}
\label{app:con_cen}
We have shown that concentration at zero occurs under the contraction condition. We now check whether the same dynamics could instead concentrate around some other deterministic charge value. The following calculation shows that it cannot. For a candidate concentration center $\chi_\star(w)$, define
its centered second moment by
\begin{equation}
  M_{2,w}(k)
  :=
  \int_{\mathbb R}
    \bigl(\chi-\chi_\star(w)\bigr)^2
    \nu_w(\dd\chi,k).
  \label{eq:app-hadamard-centered-M2}
\end{equation}
Thus concentration in second moment at $\chi_\star(w)$
means $M_{2,w}(k)\to0$. At the fixed predictor
$w$, equation~\eqref{eq:chi_exact} gives
\begin{equation}
  \chi_{k+1}-\chi_\star(w)
  =
  \bigl(1-\eta^2g_{\mathcal B}(w)^2\bigr)
  \bigl(\chi_k-\chi_\star(w)\bigr)
  -
  \eta^2g_{\mathcal B}(w)^2\chi_\star(w).
  \label{eq:app-hadamard-X-update}
\end{equation}
The new batch is independent of the charge at step $k$.
Squaring this identity and taking expectations therefore yields
\begin{align}
  M_{2,w}(k+1)
  ={}&
  \rho(w)M_{2,w}(k)
  \nonumber\\
  &-2\eta^2\chi_\star(w)
    \bigl(\mu_2(w)-\eta^2\mu_4(w)\bigr)
    \int_{\mathbb R}
      \bigl(\chi-\chi_\star(w)\bigr)
      \nu_w(\dd\chi,k)
  \nonumber\\
  &+\eta^4\mu_4(w)\chi_\star(w)^2.
  \label{eq:app-hadamard-center-recursion}
\end{align}
By Cauchy--Schwarz,
\begin{equation}
  \left|
    \int_{\mathbb R}
      \bigl(\chi-\chi_\star(w)\bigr)
      \nu_w(\dd\chi,k)
  \right|
  \leq
  \sqrt{M_{2,w}(k)}.
\end{equation}
Consequently, if $M_{2,w}(k)\to0$, the first
two terms on the right of
equation~\eqref{eq:app-hadamard-center-recursion}
vanish. Its left side also vanishes, so
\begin{equation}
  0=\eta^4\mu_4(w)\chi_\star(w)^2.
  \label{eq:app-hadamard-forced-center}
\end{equation}
For $\eta>0$ and $\mu_4(w)>0$, any deterministic
second-moment concentration center must therefore satisfy
$\chi_\star(w)=0$. At this center,
equation~\eqref{eq:app-hadamard-center-recursion}
reduces to equation~\eqref{eq:hadamard-M2-recursion}.

\section{Implicit bias on the selected charge level}
\label{app:coarea-derivation}

The previous section answered a dynamical question: at fixed predictor \(w\), which charge value survives asymptotically? This section answers a different, geometric question: once the selected value is known to be \(\chi=0\), how much parameter-space weight is associated with each predictor \(w\) on that level?\\
Let
$\Theta\subseteq\mathbb R^2$, and let
$h,\chi:\Theta\to\mathbb R$ be continuously differentiable.
Work at regular values for which the inverse images below are
discrete and their sums are well defined.

\paragraph{The level selected by charge concentration.}
Under the adiabatic assumption, fix $w$. Suppose the reduced
charge law satisfies
$$
  M_{2,w}(k)\longrightarrow 0.
$$
By equation~\eqref{eq:nu_delta}, its limiting charge
distribution is $\delta_0$. Hence the compatible parameter
values at this fixed predictor form
$$
  \mathcal S_w
  =
  \{\theta\in\Theta:h(\theta)=w,\ \chi(\theta)=0\}.
$$
If parameter laws supported on $h^{-1}(w)$ have a weakly
convergent subsequence, continuity of $h$ and $\chi$
implies that its limit is supported on $\mathcal S_w$.
This conclusion identifies the selected level, without
determining a measure on that level.

\paragraph{Coarea at the fixed predictor.}
Set $\Psi=(h,\chi)$. On a regular inverse branch,
change of variables from $\theta=(u,v)$ to
$(w,c)=(h(\theta),\chi(\theta))$ gives
$$
  \dd\theta
  =
  \frac{\dd w\,\dd c}{|\det D\Psi(\theta)|}.
$$
Summing over inverse branches and evaluating at the selected
charge $c=0$ yields, for a test function $f$,
\begin{equation}
  \int_{\Theta}
    f(\theta)\,
    \delta(h(\theta)-w)\,
    \delta(\chi(\theta))\,\dd\theta
  =
  \sum_{\theta\in\mathcal S_w}
    \frac{f(\theta)}
         {|\det D\Psi(\theta)|}.
  \label{eq:selected-coarea-identity}
\end{equation}
The delta factors denote a distribution evaluated through
this identity. They are not ordinary functions whose
pointwise product is being integrated.

Taking $f=1$, when the inverse-branch sum is finite,
gives the total geometric factor
\begin{equation}
  \Gamma_0(w)
  =
  \sum_{\theta\in\mathcal S_w}
    \frac{1}{|\det D\Psi(\theta)|}.
  \label{eq:jaco}
\end{equation}
Thus charge concentration tells us to evaluate coarea at
$\chi=0$, while change of variables supplies the
inverse-Jacobian factors. Equation~\eqref{eq:selected-coarea-identity}
specifies the resulting unnormalized geometric weight
at the fixed predictor.

The same calculation can be expressed while allowing $w$
to vary as a coordinate in the change-of-variables formula.
For a test function $f$ of the predictor,
\begin{equation}
  \int_{\Theta}
    f(h(\theta))
    \delta(\chi(\theta))\,\dd\theta
  =
  \int_{\mathbb R}
    f(w)\Gamma_0(w)\,\dd w,
  \label{eq:app-Gamma-selected}
\end{equation}
whenever both sides are defined on the regular region under
consideration. This identity says that the geometric
distribution on the selected charge level has predictor
weight $\Gamma_0(w)$. Varying $w$ here is a
change-of-variables calculation: the charge dynamics above
was analyzed separately at each fixed $w$.

Where $0<\Gamma_0(w)<\infty$, the associated geometric
bias is $L_{\mathrm{IB}}(w)=-\log\Gamma_0(w)$.
Neither this definition nor
equation~\eqref{eq:app-Gamma-selected} asserts that SGD
samples predictors according to $\Gamma_0(w)\,\dd w$.

\subsection{Hadamard implicit bias}
We now evaluate the preceding general formula for the Hadamard parametrization. The dynamics has already supplied \(\chi=0\), so only the Jacobian and the representatives in \(\mathcal S_w\) remain to be computed.\\
For $h(u,v)=uv$ and $\chi(u,v)=\tfrac12(u^2-v^2)$, the dynamically selected charge is $\chi_\star=0$. Thus,
$\Psi(u,v) = \left( uv,\frac{u^2-v^2}{2} \right)$, hence

\begin{equation}
D\Psi(u,v)
  =
  \begin{pmatrix}v&u\\u&-v\end{pmatrix},
  \qquad
  |\det D\Psi(u,v)|=u^2+v^2.
\end{equation}
Fix $w\ne0$. Solving $uv=w$ and $\chi=0$
gives exactly
$$
  (u,v)
  =
  \left(\sqrt{|w|},
        \operatorname{sgn}(w)\sqrt{|w|}\right)
  \quad\text{and}\quad
  (u,v)
  =
  \left(-\sqrt{|w|},
        -\operatorname{sgn}(w)\sqrt{|w|}\right).
$$
At both values, $|\det D\Psi|=2|w|$.
Equation~\eqref{eq:selected-coarea-identity} therefore
becomes the explicit distributional identity
\begin{equation}
\begin{aligned}
  &\int_{\mathbb R^2}
    \varphi(u,v)\,
    \delta(uv-w)\,
    \delta\!\left(\frac{u^2-v^2}{2}\right)
    \,\dd u\,\dd v\\
  &\quad=
  \frac{
    \varphi\!\left(
      \sqrt{|w|},
      \operatorname{sgn}(w)\sqrt{|w|}
    \right)
    +
    \varphi\!\left(
      -\sqrt{|w|},
      -\operatorname{sgn}(w)\sqrt{|w|}
    \right)
  }{2|w|}.
\end{aligned}
\end{equation}
In particular, its total coarea factor is
\begin{equation}
  \Gamma_0(w)
  =
  \frac{1}{|w|},
  \qquad
  L_{\mathrm{IB}}(w)=\log|w|.
\end{equation}
For $d$ independent coordinates, the joint Jacobian
is block diagonal and each coordinate has two balanced
representatives. Consequently, for $w_i\ne0$ throughout,
\begin{equation}
  \Gamma_0(w)
  =
  2^d\prod_{i=1}^{d}\frac{1}{2|w_i|}
  =
  \prod_{i=1}^{d}|w_i|^{-1},
  \qquad
  L_{\mathrm{IB}}(w)
  =
  \sum_{i=1}^{d}\log|w_i|.
\end{equation}
Thus, the geometric weight factorizes across coordinates, recovering the sparsity-promoting logarithmic bias of the Hadamard parametrization.

\section{Binary construction: geometric bias and charge dynamics}
\label{app:binary-analysis}

This appendix derives the binary construction
introduced in Section~\ref{sec:binary}. The argument has two parts. First, we inverse-design a holomorphic polynomial whose
critical points represent the targeted binary predictors and compute its
implicit bias. Second, we
study the exact discrete SGD dynamics and determine under which conditions
the parameters approach the component of the zero-charge set containing
those critical points. These two statements are kept separate: the coarea
calculation determines the geometric preference on the selected charge
level, whereas the discrete dynamics determines whether training approaches
that level.

Throughout, the target dictionary is
\(w_\star=\{-1,+1\}\).

\subsection{Inverse-designed polynomial and its critical representatives}

We seek a holomorphic polynomial \(P\) for which the charge and predictor
are defined by
\begin{equation}
    \chi(u,v)=\operatorname{Re}P(\zeta),
    \qquad
    h(u,v)=\frac12\operatorname{Im}P(\zeta),
    \qquad
    \zeta=u+\mathrm{i}v,
\end{equation}
and whose critical points represent the target values
\(w_\star=\pm1\).

For a critical point \(\zeta\), the required conditions are
\begin{equation}
    P'(\zeta)=0,
    \qquad
    \operatorname{Re}P(\zeta)=0,
    \qquad
    \frac12\operatorname{Im}P(\zeta)=w_\star.
\end{equation}
Equivalently,
\begin{equation}
    P'(\zeta)=0,
    \qquad
    P(\zeta)=2\mathrm{i}w_\star.
    \label{eq:app-binary-design-condition}
\end{equation}

We choose the two critical locations symmetrically at \(\zeta=\mathrm{i}w_\star\) for \(w_\star\in\{-1,+1\}\). The polynomial derivative of smallest degree with zeros at
\(\zeta=\pm\mathrm{i}\) is
\begin{equation}
    P'(\zeta)
    =
    a(\zeta-\mathrm{i})(\zeta+\mathrm{i})
    =
    a(\zeta^2+1),
\end{equation}
where \(a\neq0\). Integrating and choosing the additive constant equal
to zero gives
\begin{equation}
    P(\zeta)
    =
    a\left(\frac{\zeta^3}{3}+\zeta\right).
\end{equation}
At the two critical points,
\begin{equation}
    P(\mathrm{i}) =
    \frac{2a}{3}\mathrm{i}, \qquad
    P(-\mathrm{i})=
    -\frac{2a}{3}\mathrm{i}.
\end{equation}
Condition~\eqref{eq:app-binary-design-condition} therefore fixes
\(a=3\), and hence
\begin{equation}
    P'(\zeta)
    =
    3(\zeta-\mathrm{i})(\zeta+\mathrm{i}),
    \qquad
    P(\zeta)
    =
    \zeta^3+3\zeta.
    \label{eq:app-binary-polynomial}
\end{equation}

Expanding \(P\) at \(\zeta=u+\mathrm{i}v\) gives
\begin{equation}
\begin{aligned}
    P(u+\mathrm{i}v)
    &=
    (u+\mathrm{i}v)^3+3(u+\mathrm{i}v)\\
    &=
    u^3-3uv^2+3u
    +
    \mathrm{i}\left(3u^2v-v^3+3v\right).
\end{aligned}
\end{equation}
Therefore
\begin{equation}
    \chi(u,v)
    =
    u(u^2-3v^2+3),
    \qquad
    h(u,v)
    =
    \frac12v(3u^2-v^2+3).
    \label{eq:app-binary-pair}
\end{equation}
which recover equation~\eqref{eq:binary-pair}.
Differentiating gives
\begin{equation}
    \nabla h
    =
    \begin{pmatrix}
        3uv\\[2pt]
        \frac32(u^2-v^2+1)
    \end{pmatrix},
    \qquad
    \nabla\chi
    =
    \begin{pmatrix}
        3(u^2-v^2+1)\\[2pt]
        -6uv
    \end{pmatrix}.
    \label{eq:binary-gradients}
\end{equation}
Their inner product vanishes identically:
\begin{equation}
    \nabla\chi^{\mathsf T}\nabla h
    =
    9uv(u^2-v^2+1)
    -
    9uv(u^2-v^2+1)=0.
\end{equation}
Thus \(\nabla\chi\) generates a local predictor-preserving direction.

The generator vanishes when \(uv=0\) and \(u^2-v^2+1=0\). If \(u=0\), the second equation gives \(v^2=1\).
If \(v=0\), it would give \(u^2=-1\), which has no real solution.
Hence the only real fixed points are
\begin{equation}
    (u,v)=(0,+1),
    \qquad
    (u,v)=(0,-1).
\end{equation}
Equivalently, these are precisely the two critical points
\(P'(\zeta)=0\). Their predictor values are
\begin{equation}
    h(0,w_\star)=w_\star,
    \qquad
    w_\star\in\{-1,+1\}.
\end{equation}
The zero-charge set itself is larger. From
equation~\eqref{eq:app-binary-pair},
\begin{equation}
    \{\chi=0\}
    =
    \{u=0\}
    \cup
    \{u^2=3v^2-3\}.
    \label{eq:app-binary-zero-set}
\end{equation}
The designed critical representatives belong to the component
\(u=0\). This distinction will become important in the discrete
dynamical analysis below.

\subsection{Implicit bias of the binary parametrization}
We now compute the geometric weight on the zero-charge level.
Define the joint map \(\Psi(u,v)=\bigl(h(u,v),\chi(u,v)\bigr)\).
Because the binary predictor is
\(h=\operatorname{Im}P/2\), rather than
\(h=\operatorname{Im}P\), the predictor row of the joint Jacobian
carries an additional factor \(1/2\) compared with the general
holomorphic construction of Appendix~\ref{app:local-lie-symmetry}.

Using equation~\eqref{eq:binary-gradients},
\begin{equation}
    D\Psi
    =
    \begin{pmatrix}
        3uv &
        \frac32(u^2-v^2+1)\\[2pt]
        3(u^2-v^2+1) &
        -6uv
    \end{pmatrix}.
\end{equation}
Hence
\begin{equation}
    |\det D\Psi|
    =
    18u^2v^2
    +
    \frac92(u^2-v^2+1)^2.
    \label{eq:app-binary-det-direct}
\end{equation}
On the other hand,
\begin{equation}
    |P'(\zeta)|^2
    =
    9(u^2-v^2+1)^2
    +
    36u^2v^2.
\end{equation}
Comparing with equation~\eqref{eq:app-binary-det-direct} gives
\begin{equation}
    |\det D\Psi|
    =
    \frac12|P'(\zeta)|^2.
    \label{eq:app-binary-jacobian}
\end{equation}
At a fixed predictor value \(w\), imposing simultaneously
\(\chi=0\) and \(h=w\) is equivalent to \(P(\zeta) = \chi(\zeta)+2\mathrm{i}h(\zeta) = 2\mathrm{i}w\). Thus the selected set is
\begin{equation}
    \mathcal S_w
    =
    \{\zeta\in\mathbb C:
      P(\zeta)=2\mathrm{i}w\}.
\end{equation}
For a regular predictor value, the geometric weight is therefore
\begin{equation}
    \Gamma_0(w)
    =
    \sum_{\zeta:\,P(\zeta)=2\mathrm{i}w}
    \frac{1}{|\det D\Psi(\zeta)|}
    =
    \sum_{\zeta:\,P(\zeta)=2\mathrm{i}w}
    \frac{2}{|P'(\zeta)|^2}.
    \label{eq:app-binary-full-coarea}
\end{equation}
This is a geometric weight on the zero-charge level.

We next derive its behaviour near the targeted binary
values. Fix \(w_\star\), with corresponding critical
point \(\zeta=\mathrm{i}w_\star\). At the target value itself, the polynomial factors exactly as
\begin{equation}
    P(\zeta)-2\mathrm{i}w_\star
    =
    (\zeta-\mathrm{i}w_\star)^2
    (\zeta+2\mathrm{i}w_\star).
    \label{eq:app-binary-target-factorization}
\end{equation}
Thus the designed critical representative
\(\zeta=\mathrm{i}w_\star\) is a double root, while
\(\zeta=-2\mathrm{i}w_\star\) is a distinct regular
representative of the same predictor.

To determine the behaviour for nearby regular values, write \(\zeta =\mathrm{i}w_\star+\xi\). Since \(P'(\mathrm{i}w_\star)=0\), direct expansion gives
\begin{equation}
\begin{aligned}
    P(\mathrm{i}w_\star+\xi)
    &=
    2\mathrm{i}w_\star
    +
    3\mathrm{i}w_\star\xi^2
    +
    \xi^3.
\end{aligned}
\label{eq:binary-critical-expansion}
\end{equation}
Hence
\(P(\zeta)=2\mathrm{i}w\) is equivalent to
\begin{equation}
    \xi^2
    \left(
        3\mathrm{i}w_\star+\xi
    \right)
    =
    2\mathrm{i}(w-w_\star).
    \label{eq:app-binary-local-equation}
\end{equation}
Two roots approach \(\xi=0\) as \(w\to w_\star\).
For either of these roots,
\begin{equation}
    |\xi|^2=
    \frac{
        2|w-w_\star|
    }{
        |3\mathrm{i}w_\star+\xi|
    } =
    \frac23
    |w-w_\star|
    \bigl(1+o(1)\bigr).
    \label{eq:app-binary-xi-asymptotic}
\end{equation}

Moreover,
\begin{align}
    P'(\mathrm{i}w_\star+\xi)
    &=
    6\mathrm{i}w_\star\xi
    +
    3\xi^2
    \nonumber\\
    &=
    3\xi
    \left(
        2\mathrm{i}w_\star+\xi
    \right).
\end{align}
Therefore, using
equation~\eqref{eq:app-binary-xi-asymptotic},
\begin{equation}
\begin{aligned}
    |P'(\mathrm{i}w_\star+\xi)|^2
    &=
    9|\xi|^2
    |2\mathrm{i}w_\star+\xi|^2\\
    &=
    24|w-w_\star|
    \bigl(1+o(1)\bigr).
\end{aligned}
\end{equation}
Equation~\eqref{eq:app-binary-jacobian} then yields
\begin{equation}
    |\det D\Psi|
    =
    12|w-w_\star|
    \bigl(1+o(1)\bigr).
\end{equation}
Consequently, each of the two representatives approaching the
critical point contributes
\begin{equation}
    \frac{1}{
        12|w-w_\star|
    }
    \bigl(1+o(1)\bigr)
\end{equation}
to \(\Gamma_0(w)\).

The third representative approaches
\(\zeta=-2\mathrm{i}w_\star\). Since \(P'(-2\mathrm{i}w_\star)=-9\), its contribution remains finite.
The two singular contributions therefore dominate, giving
\begin{equation}
    \Gamma_0(w)
    =
    \frac{1+o(1)}{
        6|w-w_\star|
    },
    \qquad
    w\to w_\star.
    \label{eq:app-binary-local-gamma}
\end{equation}
Accordingly,
\begin{align}
    L_{\mathrm{IB}}(w)
    &=
    -\log\Gamma_0(w)
    \nonumber\\
    &=
    \log|w-w_\star|
    +
    \log6
    +
    o(1).
\label{eq:app-binary-local-bias}
\end{align}
Thus the geometric weight diverges as the predictor approaches either
targeted binary value. The inverse-Jacobian formula is used only for
regular values approaching \(w_\star\). At \(w=w_\star\), the
Jacobian vanishes at the critical representative and its inverse is
not defined there. Moreover, the singular factor
\(1/|w-w_\star|\) is not locally integrable, so
equation~\eqref{eq:app-binary-local-gamma} should be interpreted as
the local behaviour of the geometric coarea weight, not as a normalized
predictor probability density.

\subsection{Exact finite-step charge dynamics}

We now turn to discrete SGD. For minibatch \(\mathcal B_k\), define \(g_{\mathcal B} \equiv g_{\mathcal B_k} = \partial_w \ell_{\mathcal B_k} \bigl(h(u_k,v_k)\bigr)\). Since
\begin{equation}
    \nabla h(u_k,v_k)
    =
    \begin{pmatrix}
        3u_kv_k\\[2pt]
        \frac32(u_k^2-v_k^2+1)
    \end{pmatrix},
\end{equation}
To derive the charge change, define the parameter increments
\begin{equation}
    \Delta u_k
    :=
    u_{k+1}-u_k
    =
    -3\eta g_{\mathcal B}u_kv_k,
    \qquad
    \Delta v_k
    :=
    v_{k+1}-v_k
    =
    -\frac32\eta g_{\mathcal B}
    (u_k^2-v_k^2+1).
    \label{eq:app-binary-increments}
\end{equation}
Since \(\chi(u,v)=u^3-3uv^2+3u\)
is a cubic polynomial, its Taylor expansion around
\((u_k,v_k)\) is exact at third order. Writing
\begin{equation}
    A_k:=u_k^2-v_k^2+1,
\end{equation}
the required derivatives are
\begin{align}
    \partial_u\chi(u_k,v_k)
    &=
    3A_k,
    \qquad
    \partial_v\chi(u_k,v_k)
    =
    -6u_kv_k,
    \nonumber\\
    \partial_{uu}\chi(u_k,v_k)
    &=
    6u_k,
    \qquad
    \partial_{uv}\chi(u_k,v_k)
    =
    -6v_k,
    \nonumber\\
    \partial_{vv}\chi(u_k,v_k)
    &=
    -6u_k, \qquad \partial_{uuu}\chi=6,
    \nonumber\\
    \partial_{uvv}\chi=-6.
\end{align}
Therefore the exact expansion is
\begin{align}
    \chi_{k+1}-\chi_k
    ={}&
    3A_k\Delta u_k
    -6u_kv_k\Delta v_k
    \nonumber\\
    &+
    3u_k(\Delta u_k)^2
    -6v_k\Delta u_k\Delta v_k
    -3u_k(\Delta v_k)^2
    \nonumber\\
    &+
    (\Delta u_k)^3
    -3\Delta u_k(\Delta v_k)^2.
    \label{eq:app-binary-explicit-taylor}
\end{align}
The first-order contribution is
\begin{align}
    3A_k\Delta u_k
    -6u_kv_k\Delta v_k
    &=
    3A_k
    \left(
        -3\eta g_{\mathcal B}u_kv_k
    \right)
    \nonumber\\
    &\quad
    -6u_kv_k
    \left(
        -\frac32\eta g_{\mathcal B}A_k
    \right)
    \nonumber\\
    &=
    -9\eta g_{\mathcal B}u_kv_kA_k
    +
    9\eta g_{\mathcal B}u_kv_kA_k
    \nonumber\\
    &=0.
\label{eq:app-binary-first-order}
\end{align}
Thus the complete \(O(\eta)\) contribution cancels exactly, as
required by the predictor-preserving symmetry.
We next compute the quadratic contribution. From
equation~\eqref{eq:app-binary-increments},
\begin{align}
    (\Delta u_k)^2
    &=
    9\eta^2g_{\mathcal B}^2u_k^2v_k^2,
    \nonumber\\
    \Delta u_k\Delta v_k
    &=
    \frac92
    \eta^2g_{\mathcal B}^2u_kv_kA_k,
    \nonumber\\
    (\Delta v_k)^2
    &=
    \frac94
    \eta^2g_{\mathcal B}^2A_k^2.
\end{align}
Hence
\begin{align}
    &
    3u_k(\Delta u_k)^2
    -6v_k\Delta u_k\Delta v_k
    -3u_k(\Delta v_k)^2
    \nonumber\\
    &=
    27\eta^2g_{\mathcal B}^2u_k^3v_k^2
    -
    27\eta^2g_{\mathcal B}^2u_kv_k^2A_k
    -
    \frac{27}{4}
    \eta^2g_{\mathcal B}^2u_kA_k^2
    \nonumber\\
    &=
    \frac{27}{4}
    \eta^2g_{\mathcal B}^2u_k
    \left[
        4u_k^2v_k^2
        -
        4v_k^2A_k
        -
        A_k^2
    \right].
\end{align}
Since \(4u_k^2v_k^2-4v_k^2A_k=4v_k^2(v_k^2-1)\),
the quadratic term reduces to
\begin{equation}
    \frac{27}{4}
    \eta^2g_{\mathcal B}^2u_k
    \left[
        4v_k^2(v_k^2-1)
        -
        A_k^2
    \right].
    \label{eq:app-binary-second-order}
\end{equation}
The cubic contribution is
\begin{align}
    (\Delta u_k)^3
    -3\Delta u_k(\Delta v_k)^2
    ={}&
    -27\eta^3g_{\mathcal B}^3u_k^3v_k^3
    \nonumber\\
    &+
    \frac{81}{4}
    \eta^3g_{\mathcal B}^3u_kv_kA_k^2
    \nonumber\\
    ={}&
    -\frac{27}{4}
    \eta^3g_{\mathcal B}^3u_kv_k
    \left[
        4u_k^2v_k^2
        -
        3A_k^2
    \right].
\label{eq:app-binary-third-order}
\end{align}
Combining equations~\eqref{eq:app-binary-first-order},
\eqref{eq:app-binary-second-order}, and
\eqref{eq:app-binary-third-order} gives the exact finite-step
charge update
\begin{align}
    \chi_{k+1}-\chi_k
    ={}&
    \frac{27}{4}
    \eta^2g_{\mathcal B}^2u_k
    \left[
        4v_k^2(v_k^2-1)
        -
        (u_k^2-v_k^2+1)^2
    \right]
    \nonumber\\
    &-
    \frac{27}{4}
    \eta^3g_{\mathcal B}^3u_kv_k
    \left[
        4u_k^2v_k^2
        -
        3(u_k^2-v_k^2+1)^2
    \right].
\label{eq:binary-direct-charge-update}
\end{align}
Thus the predictor-preserving symmetry cancels the complete
first-order change in the charge. Because \(\chi\) is cubic, the
remaining terms terminate exactly at orders \(\eta^2\) and
\(\eta^3\).

\subsection{A sufficient condition for attraction to the invariant branch}
We now derive a sufficient condition under which a trajectory starting
with \(u_0\neq0\) approaches the invariant branch \(u=0\).

From equation~\eqref{eq:app-binary-increments}, \(u_{k+1}=(1-3\eta v_k g_{\mathcal B})u_k\).
Assume that \(1-3\eta v_k g_{\mathcal B}\neq0\), and suppose that, conditionally on the current parameters
\((u_k,v_k)\), there exists a constant \(\varepsilon>0\) such that
\begin{equation}
    \mathbb E_{\mathcal B}
    \left[
        1-3\eta v_k g_{\mathcal B}
        \mid
        u_k,v_k
    \right]
    \leq
    -\varepsilon
    \qquad
    \text{for every }k.
    \label{eq:app-binary-log-condition}
\end{equation}
Assume additionally that the corresponding conditional second moments
are uniformly bounded,
\begin{equation}
    \sup_k
    \mathbb E_{\mathcal B}
    \left[
        (1-3\eta v_k g_{\mathcal B})^2
        \mid
        u_k,v_k
    \right]
    <\infty.
    \label{eq:app-binary-log-second-moment}
\end{equation}
Iterating the multiplicative update gives the exact identity
\begin{equation}
    u_N
    =
    u_0
    \prod_{k=0}^{N-1}
    \left(
        1-3\eta v_k g_{\mathcal B}
    \right),
\end{equation}
and therefore
\begin{equation}
    \log\frac{|u_N|}{|u_0|}
    =
    \sum_{k=0}^{N-1}
    \log\left|1-3\eta v_k g_{\mathcal B}\right|.
    \label{eq:app-binary-log-sum}
\end{equation}
We decompose each logarithmic increment into its conditional mean and
its zero-mean fluctuation,
\begin{align}
    \log\left|1-3\eta v_k g_{\mathcal B}\right|
    ={}&
    \mathbb E_{\mathcal B}
    \left[
        \log\left|1-3\eta v_k g_{\mathcal B}\right|
        \mid
        u_k,v_k
    \right]
    \nonumber\\
    &+
    \left(
        \log\left|1-3\eta v_k g_{\mathcal B}\right|
        -
        \mathbb E_{\mathcal B}
        \left[
            \log\left|1-3\eta v_k g_{\mathcal B}\right|
            \mid
            u_k,v_k
        \right]
    \right).
\end{align}
Because a fresh minibatch is sampled at each step, the terms in
parentheses are martingale differences. Under the uniform
second-moment bound in
equation~\eqref{eq:app-binary-log-second-moment}, their empirical
average converges to zero:
\begin{equation}
    \frac1N
    \sum_{k=0}^{N-1}
    \left(
        \log\left|1-3\eta v_k g_{\mathcal B}\right|
        -
        \mathbb E_{\mathcal B}
        \left[
            \log\left|1-3\eta v_k g_{\mathcal B}\right|
            \mid
            u_k,v_k
        \right]
    \right)
    \longrightarrow0.
\end{equation}
Hence, dividing equation~\eqref{eq:app-binary-log-sum} by \(N\),
\begin{align}
    \limsup_{N\to\infty}
    \frac1N
    \log\frac{|u_N|}{|u_0|}
    &=
    \limsup_{N\to\infty}
    \frac1N
    \sum_{k=0}^{N-1}
    \mathbb E_{\mathcal B}
    \left[
        \log\left|1-3\eta v_k g_{\mathcal B}\right|
        \mid
        u_k,v_k
    \right]
    \nonumber\\
    &\leq
    -\varepsilon.
\label{eq:binary-branch-convergence}
\end{align}
Therefore \(u_N\longrightarrow0\). If \(v_N\) remains bounded, then \(|\chi_N|\to0\). Thus attraction to the invariant branch \(u=0\) also implies
convergence of the charge to zero, provided \(v_N\) remains bounded.

\subsection{Dynamical selection of the critical representatives}
The previous subsection establishes, under the stated conditions,
attraction toward \(u=0\). This is enough to imply
\(\chi_k\to0\) when \(v_k\) remains bounded, but it is not enough by
itself to determine the limiting value of \(v_k\).

On the invariant branch \(u=0\), the predictor and the SGD update reduce to
\begin{equation}
    h(0,v)
    =
    \frac12(-v^3+3v),
    \qquad
    v_{k+1}
    =
    v_k
    -
    \frac32\eta g_{\mathcal B}(1-v_k^2).
    \label{eq:app-binary-branch-dynamics}
\end{equation}
The factor \(1-v_k^2\) already distinguishes the two values
\(v=\pm1\): at these points the parameter update vanishes for every
sampled gradient. Moreover, \(h(0,\pm1)=\pm1\). Thus the designed binary representatives are precisely the points on
the invariant branch where the remaining SGD motion in \(v\) stops.

Suppose now that the predictor converges to a targeted value \(h(u_k,v_k)\longrightarrow w_\star\), and that \(v_k\) remains bounded. Every accumulation point then
satisfies
\begin{equation}
    u=0,
    \qquad
    h(0,v)=w_\star.
\end{equation}
The latter equation factors exactly as
\begin{equation}
\begin{aligned}
    h(0,v)-w_\star
    &=
    \frac{-v^3+3v}{2}
    -
    w_\star\\
    &=
    -\frac12
    (v-w_\star)^2
    (v+2w_\star).
\end{aligned}
\label{eq:binary-alternative-representatives}
\end{equation}
Thus there are exactly two distinct compatible parameter
representatives:
\begin{equation}
    (u,v)
    =
    (0,w_\star),
    \qquad
    (u,v)
    =
    (0,-2w_\star).
    \label{eq:app-binary-two-representatives}
\end{equation}
These are exactly the two distinct points already identified by the
factorization
\begin{equation}
    P(\zeta)-2\mathrm{i}w_\star
    =
    (\zeta-\mathrm{i}w_\star)^2
    (\zeta+2\mathrm{i}w_\star).
\end{equation}
The first is the designed critical representative and occurs as a
double root. The second is the regular representative that has a finite
contribution equal to \(2/81\).

At the designed representative, \(v=w_\star\) and \(1-v^2=0\), equation~\eqref{eq:app-binary-branch-dynamics} gives \(v_{k+1}=v_k=w_\star\),
for every sampled gradient. Equivalently,
\begin{equation}
    \nabla h(0,w_\star)=0.
\end{equation}
Hence the critical representative is absorbing under SGD.

At the alternative representative \((0,-2w_\star)\), instead,
\(\partial_v h=-9/2\neq0\). Writing \(v_{k+1}=-2w_\star+\delta_k\) where \(\delta_k=\frac92\eta g_{\mathcal B}\), the predictor after one step is
\begin{equation}
    h_{k+1}-w_\star
    =
    -\frac92\delta_k
    +3w_\star\delta_k^2
    -\frac12\delta_k^3.
\end{equation}
Hence a generic nonzero sampled gradient moves the predictor away
from \(w_\star\). The critical representative is therefore the point
at which the binary predictor value is preserved under the stochastic
dynamics, whereas the regular representative produces the same
predictor only instantaneously.

\begin{figure}[t]
    \centering
    \includegraphics[width=0.48\linewidth]
    {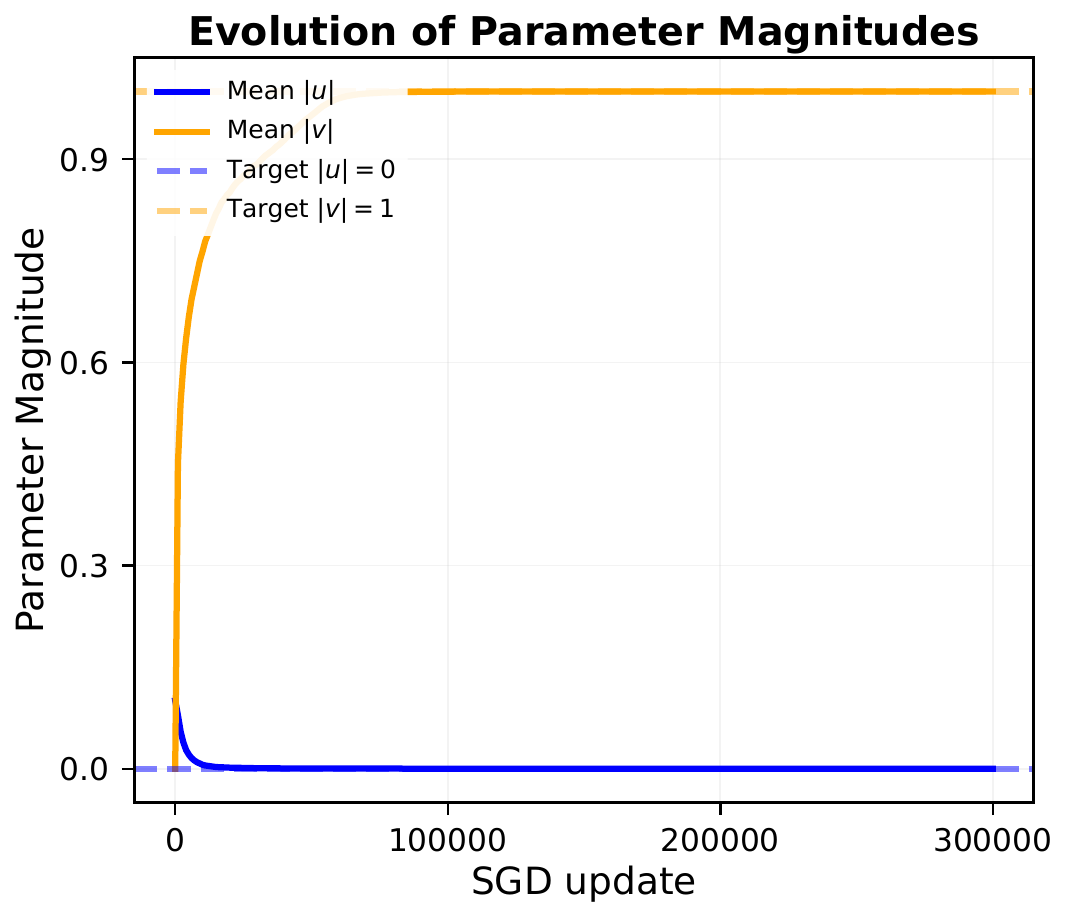}
    \hfill
    \includegraphics[width=0.48\linewidth]
    {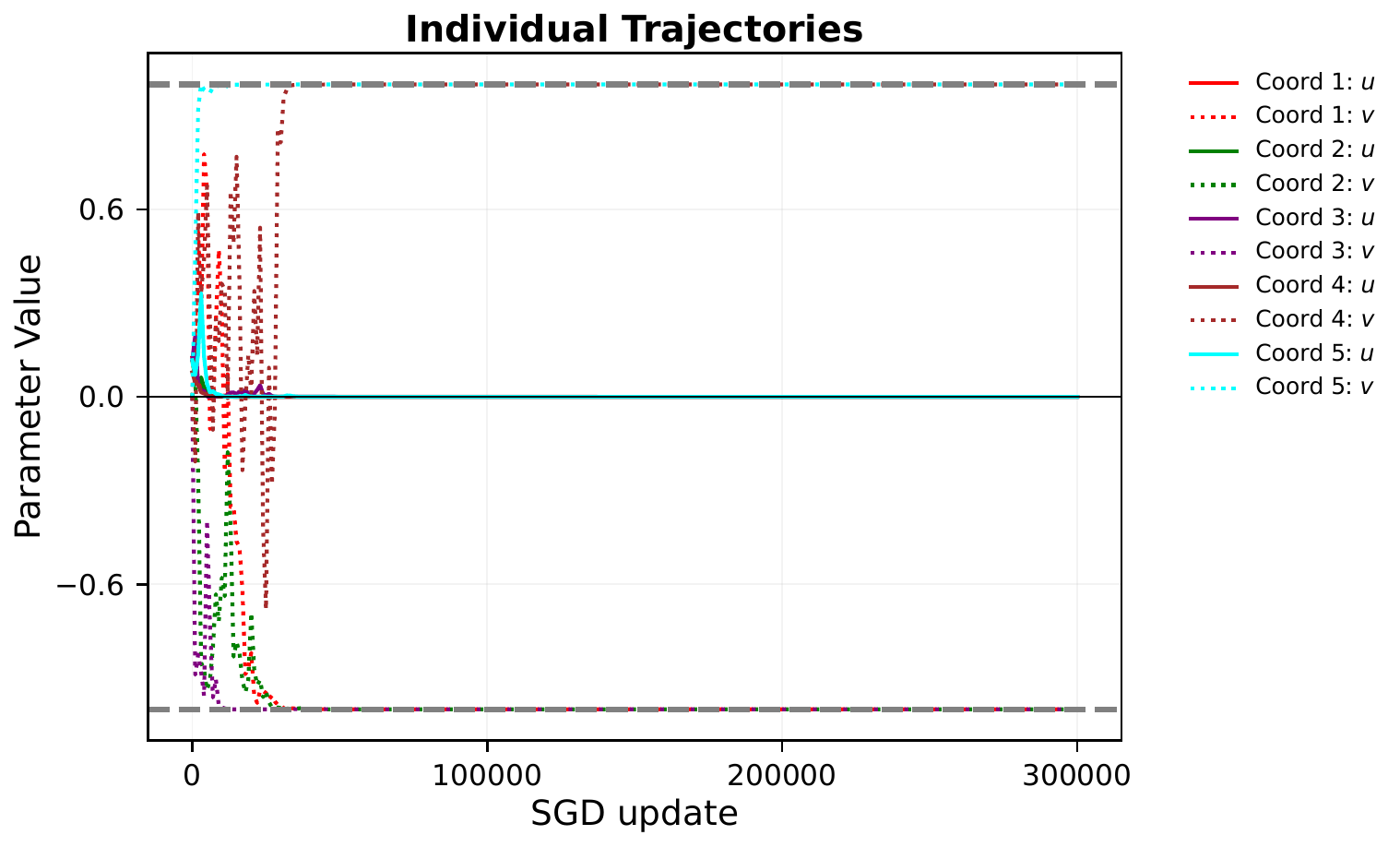}

    \caption{
    \textbf{Parameter dynamics of the parameterized model.}
    \textbf{Left:} evolution of the mean parameter magnitudes during training.
    The \(u\)-parameters approach zero while the magnitude of the
    \(v\)-parameters approaches one.
    \textbf{Right:} five individual parameter trajectories from the first run. The \(u\)-coordinates approach zero, while
    the corresponding \(v\)-coordinates approach either \(+1\) or
    \(-1\). These trajectories are consistent with attraction toward
    the invariant branch \(u=0\) and stabilization at the critical
    representatives \((0,\pm1)\) derived above.
    }
    \label{fig:binary-parameter-trajectories}
\end{figure}

\end{document}